\documentclass[review,number,sort&compress]{elsarticle}

\usepackage{amsmath}
\usepackage{algorithmic}
\usepackage{algorithm2e}
\usepackage{multirow}

\usepackage{pifont}    
\usepackage{natbib}    
\usepackage{geometry}  
\usepackage{fleqn}     
\usepackage{graphicx}  
\usepackage{txfonts}   
\usepackage{hyperref}  

\usepackage{subfigure}

\begin{document}

\title{Multi-objective Optimization of Energy Consumption and Execution Time in a Single Level Cache Memory for Embedded Systems}

\author[uex]{Josefa D\'{i}az \'{A}lvarez}
\ead{mjdiaz@unex.es}
\author[ucm]{Jos\'{e} L. Risco-Mart\'{i}n\corref{cor1}}
\ead{jlrisco@ucm.es}
\author[urjc]{J. Manuel Colmenar}
\ead{josemanuel.colmenar@urjc.es}

\cortext[cor1]{Corresponding author}
\address[uex]{Centro Universitario de M\'{e}rida, Universidad de Extremadura, 06800 M\'{e}rida, Spain}
\address[ucm]{Dpt. of Computer Architecture and Automation, Complutense University of Madrid, C/Prof. Jos\'{e} Garc\'{i}a Santesmases 9, 28040 Madrid, Spain}
\address[urjc]{Dpt. of Computer Science, Rey Juan Carlos University, 28933 M\'{o}stoles, Spain}

\begin{abstract}
Current embedded systems are specifically designed to run multimedia applications. These applications have a big impact on both performance and energy consumption. Both metrics can be optimized selecting the best cache configuration for a target set of applications. Multi-objective optimization may help to minimize both conflicting metrics in an independent manner. In this work, we propose an optimization method that based on Multi-Objective Evolutionary Algorithms, is able to find the best cache configuration for a given set of applications. To evaluate the goodness of candidate solutions, the execution of the optimization algorithm is combined with a static profiling methodology using several well-known simulation tools. Results show that our optimization framework is able to obtain an optimized cache for Mediabench applications. Compared to a baseline cache memory, our design method reaches an average improvement of 64.43\% and 91.69\% in execution time and energy consumption, respectively.
\end{abstract}


\maketitle

\section{Introduction}

Multimedia embedded systems like digital cameras, audio and video players, smartphones, etc., are one of the major driving forces in technology. Currently, they have less powerful resources than desktop systems, but these systems must run multimedia software (video, audio, gaming, etc.). These applications require high performance and consume much energy, which reduces the battery lifetime. The battery in embedded systems is limited in capacity and size because of design constraints. Hence, embedded systems designers must be very concerned on both increasing performance and reducing energy consumption, which in turn will also affect to the lifetime of the device.

In recent years, a number of scientific papers have been published indicating that the memory subsystem acts as an energy bottleneck of the system~\cite{Kandemir2006}. In fact, cache memory behavior affects both performance and energy consumption. The best cache configuration gives us the minimum execution time and the lowest energy consumption. Total cache and block sizes, associativity, and algorithms for search, prefetch and replacement, or write policies are some of the parameters that form a cache configuration. Finding optimal values for these parameters will guide us to reach the best performance and energy consumption. Finding an optimal cache configuration for one single application is a bad choice for other applications with different memory access patterns \cite{Hennessy2011}. Thus, we tackle the problem of finding the optimal cache configuration for all the applications executed in an embedded device, which will improve performance and energy consumption. 

Energy optimization directly affects aging of transistors, which is a limiting factor for long term reliability of devices. In a common context where the lifetime of a device is determined by the earliest failing component, the aging impact is more serious on memory arrays, where failure of a single SRAM cell would cause the failure of the whole system. Previous works have shown that saving energy in the memory subsystem can effectively control aging effects and can extend lifetime of the cache significantly \cite{Cai2006}, \cite{Mahmood2014}. Our approach, which optimizes performance and energy, is also indirectly improving the long term reliability of the target device. 

A first brute-force approach to obtain the best cache configuration would require the execution and evaluation of time and energy for all available cache configurations and target applications, which is an unmanageable task given current time-to-market reduced windows. In addition, execution time and energy consumption are conflicting objectives in practice. For example, if associativity is increased, the number of misses is reduced, as well as the execution time. However, a high associativity increases the hardware complexity and thus the energy consumed by the cache memory \cite{Hennessy2011}. Therefore, we present in this paper a new methodology to evaluate cache configurations in order to customize cache designs with the aim of reducing both the execution time and the energy consumption by means of a multi-objective optimization \cite{Deb2009}. In particular, our optimization framework is built around the Non-dominated Sorting Genetic Algorithm II (NSGA-II) \cite{Deb2002}. In order to evaluate our approach, we have automatically designed caches optimized for a set of multimedia applications taken from Mediabench benchmarks~\cite{mediabench}, since they are representative for image, audio and video processing. Our hardware architecture is based on the ARM920T processor \cite{ARM}, broadly used in multimedia embedded devices.

The rest of the paper is organized as follows. Next Section summarizes the related work on the topic. Section~\ref{sec:dspace} describes the design of the search space for our multi-objective optimization. Section~\ref{sec:multi} shows details of our multi-objective function, describing both performance and energy models. Our optimization framework is integrated and explained in Section ~\ref{sec:Methodology}. Then, Section \ref{sec:Experiments} analyzes our experimental results. In Section ~\ref{sec:Conclusions}, we present our conclusions based on the results obtained, and explain the main lines of our future work.

\section{Related work}

The optimization of performance and energy consumption in the memory subsystem have received a lot of attention in the last decade. Regarding performance, multiple research works have been developed with the aim of improving performance through changing architectural parameters. With respect to energy, previous studies have demonstrated that half of the energy consumption in embedded systems is due to the cache memory~\cite{Kandemir2006}. The optimization of these parameters has been conducted mainly using two different techniques: dynamic reconfiguration and static profiling.


Regarding dynamic reconfiguration, Givargis~ \cite{Givargis2006} improved cache performance by choosing a variable set of bits used as index into the cache. Zhang minimized the energy consumption introducing a new cache design method called way concatenation to reconfigure the cache by software \cite{Zang2013}. However, this approach provided a limited number of cache configurations, allowing the system engineer to optimize associativity (one-way, two-way or four-way), cache size and line size. Chen and Zou~\cite{Chen-2007} proposed an efficient reconfiguration management algorithm to optimize three parameters: cache size, line size and associativity. Similarly, Gordon-Ross and Frank Vahid~\cite{Gordon-2007} presented a dynamic tuning methodology to optimize cache sizes (2, 4, or 8 KBytes), line sizes (16, 32, or 64 bytes), and associativity (1-way, 2-way or 4-way). L\'{o}pez et al.~\cite{Sonia3} proposed an on-line algorithm on \emph{Simultaneous Multithreading (SMT)} to decide the cache configuration after a fixed set of instructions, a technique based on a cache working-set adaptation~\cite{Gracia2014}. Dynamic reconfiguration in soft real-time embedded systems on single-level cache hierarchy was proposed by Wang et al. in~\cite{Wang-2009}, and on a multi-level cache hierarchy in~\cite{Wang2011}. More recently, Wang et al.~\cite{Wang2012} minimize energy consumption in real-time embedded systems performing dynamic analysis during runtime. All these approaches optimize cache size (1, 2 or 4KB), line-size (16, 32 or 64 bytes) and associativity (1-way, 2-way or 4-way). The main inconvenient of dynamic reconfiguration is the addition of extra complexity in the design of the memory subsystem. We also see that these approaches only optimize a few number of cache parameters, minimizing either execution time or energy consumption. In addition, it is proved in this work that an offline multi-objective optimization may find optimal cache parameter values, without the need of adding hardware complexity to the standard memory subsystem design.


With respect to the use of static profiling, Rackesh Reddy in ~\cite{Reddy2010} studied the effect of multiprogramming workloads on the data cache in a preemptive multitasking environment, and proposed a technique that mapping tasks to different cache partitions, significantly reduced both dynamic and leakage power. Our approach is different, since we try to obtain the behavior of a target set of applications, obtaining their full static profile and the best memory cache configuration (i.e., size, associativity, and replacement and prefetching algorithms for both data and instruction caches) for the whole set. Andrade et al. presented in \cite{Andrade2007} an extension of a systematic analytical modeling technique based on probabilistic miss equations, allowing the automated analysis of the cache behavior for codes with irregular access patterns resulting from indirections. Nevertheless, these models can only optimize cache size and associativity. Feng et al.~\cite{Feng2011} applied  a new cache replacement policy to perform the replacement decision based on the reuse information of the cache lines and the requested data developing two reuse information predictors: a profile-based static predictor and a runtime predictor. Similarly, Xingyan and Hongyan~\cite{Xingyan2010}, based on a profiling scheme of the OPT cache replacement, presented a method to generate best static cache hints. However, these approaches only improve the replacement algorithm. Gordon-Ross et al. \cite{Gordon-Ross-2012} studied the interaction of code reordering and cache configuration, obtaining excellent results. However, this technique is applied to the intruction cache, and our systematic optimization method is applied to the full configuration of both the instruction and data caches.


Additionally, all the aforementioned approaches minimized either execution time or energy consumption. We propose the use of multi-objective optimization to simultaneously minimize both objectives. To this end, we use the concept of multi-objective optimization, which can be easily applied in evolutionary computation. Evolutionary computation and multi-objective optimization are being widely used in \emph{Computer Aided Design (CAD)} problems. Close to cache optimization, Risco et al.~\cite{Risco-2008} applied a novel parallel multi-objective evolutionary algorithm to optimize desktop applications for their use in multimedia embedded systems, improving performance, memory usage and energy consumption of the memory subsystem. In ~\cite{JDiaz-2009} a simple online \emph{Genetic Algorithm (GA)} was used to obtain the best cache associativity to improve the performance of SMT processors. In this line, Bui et al. ~\cite{Bui-2008} proposed a solution for the cache interference problem applying cache partitioning techniques using a simple GA whose solution sets the size of each cache partition and assigns tasks to partitions such that system worst-case utilization is minimized thus increasing real-time schedulability. An approach based on NSGA-II algorithm was used in~\cite{Filho-2008} to evaluate cache configurations on a second cache-level in order to improve energy consumption and performance, optimizing cache size, line size and associativity. However, none of these approaches is able to simultaneously optimize cache performance and energy consumption for a target set of applications as our methodology performs.

To the best of our knowledge none of the previous works tackle the optimization of all the parameters that we propose in this research work. Most of the cited papers focus their space exploration on cache size, line size and associativity, even though the possible values for each configurable parameter is quite small. In this work, we optimize the following cache parameters: cache size, line size, associativity, replacement policy, prefetch policy and write policy. We also consider first-level (L1) instruction/data caches, although the methodology proposed can be applied to other cache types. All the aforementioned configurable parameters complete the chromosome in the \emph{Multi-Objective Evolutionary Algorithm (MOEA)} proposed. The aim is to find the best cache configurations that minimizes memory access time (performance) and energy consumption. As we try to minimize two conflicting objectives, multi-objective optimization is suitable to address this problem. Our approach is valid on embedded systems, where the small number of applications allows the engineer to select one cache design among all the optimizations performed, as we show in this work.

\section{Design of the search space representation}\label{sec:dspace}

\begin{figure*}[ht]
	\centering
 	\includegraphics[width=0.75\textwidth]{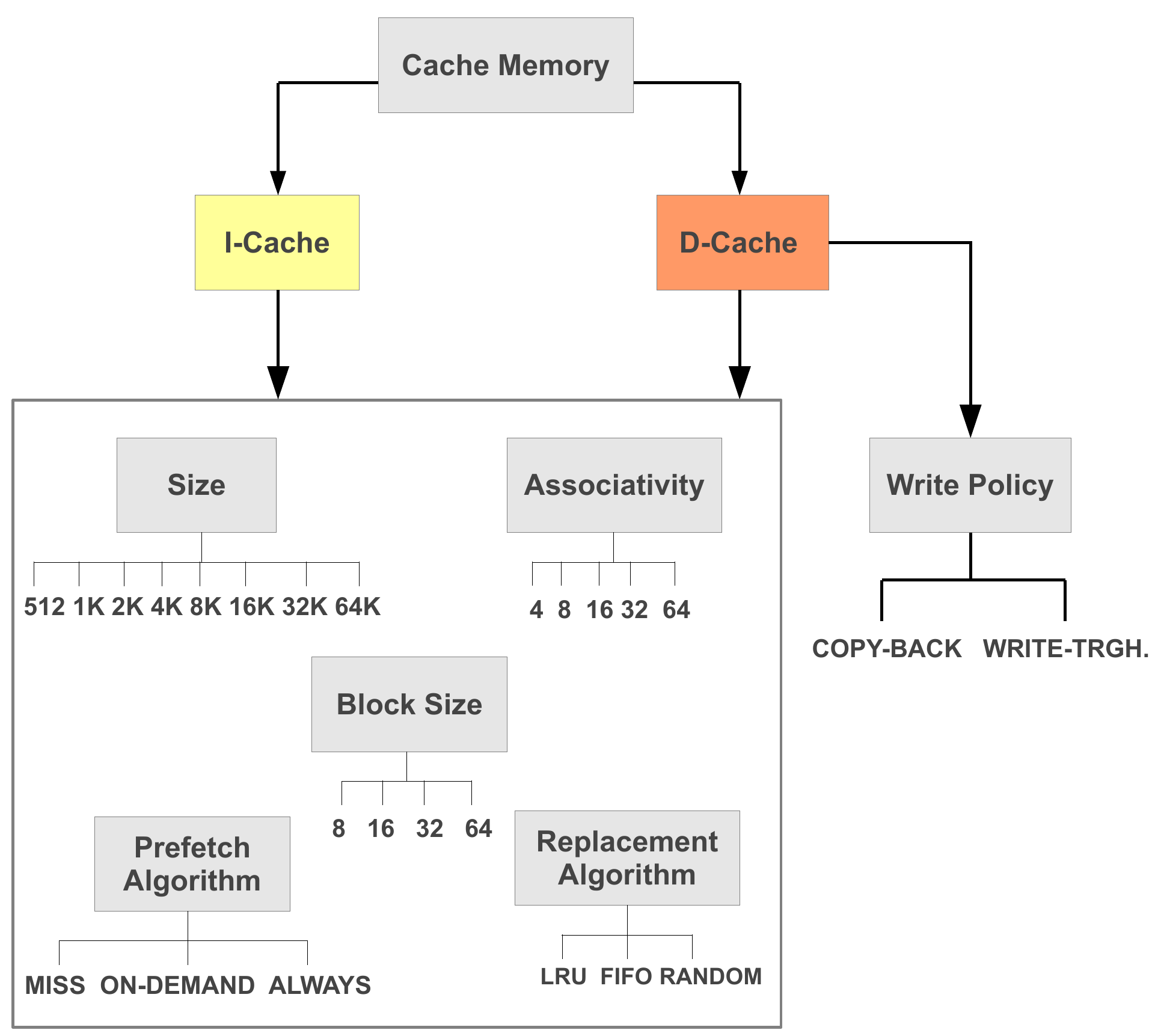}
 	\caption{Taxonomy for a cache configuration. Both instruction and data caches, labeled as I-Cache and D-Cache, must be customized with available values.}
  \label{figure:arboles} 
\end{figure*}

Designing a cache memory implies the configuration of the set of parameters that define it: cache size, line size, replacement algorithm, associativity and prefetch algorithm for both the instruction and data caches, and also write policy for data cache. Figure \ref{figure:arboles} shows these parameters and the possible candidate values that we consider in our research for the instruction cache (labeled as I-Cache) and data cache (labeled as D-Cache). These possible values, most of them illustrated in Figure \ref{figure:arboles}, are described below:

\begin{itemize}
 \item Cache size: memory cache capacity in bytes. We consider a fixed cache size of 16 KB, which is the default size of the ARM920T cache~\cite{ARM}, our target system.
 \item Line size: cache memory is divided into lines (blocks). When a miss takes place, a whole line is moved from main memory to cache memory. Possible values for this parameter are 8, 16, 32 and 64 bytes.
 \item Cache replacement algorithm: set of techniques designed to replace blocks. Algorithms selected to evaluate are: \emph{Last Recently Used (LRU)}, \emph{First Input First Output (FIFO)} and RANDOM algorithms.
 \item Associativity: the degree of associativity refers to the number of places in a cache where a block can be located. It is defined by the number of ways. In this work we deal with 4, 8, 16, 32 and 64 ways.
 \item Prefetch algorithm: determines the policy to carry blocks to the cache memory. We consider three of them: (1) when a cache miss occurs (MISS-PREFETCH), (2) only when the data is required (ON-DEMAND), and (3) when data from a block is referenced, the following block is also prefetched (ALWAYS-PREFETCH). 
  \item Write policy: it is designed to keep consistency between cache and main memory when data is modified in cache memory. Data stored in L1-cache can be written to main memory only when absolutely needed, with a COPY-BACK policy, or maybe written to cache and main memory simultaneously, with a WRITE-TROUGH policy.  
\end{itemize}

Hence, the size of the search space is 64800 cache configurations for each cache size. Thereby, a deterministic technique can take huge time slots to find an optimal solution (more than four months, as shown in Section \ref{sec:Experiments}), since each configuration must be evaluated with a program trace. In this regard, heuristic techniques fit well in solving the multi-objective optimization problem especially when a set of conflicting design objectives must be minimized. MOEAs usually provide good results in a multi-objective environment. In this context, a set of candidate solutions called individuals evolves improving a multi-objective fitness function (an individual is formed by a chromosome plus the associated value of the multi-objective function). 

\begin{figure*}
  \centering
  \includegraphics[width=0.95\textwidth]{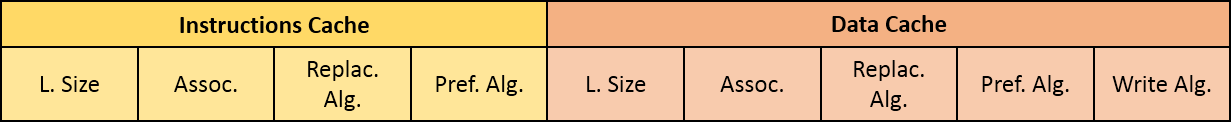}
  \caption{Chromosome Specification.}
  \label{fig:Chromosome} 
\end{figure*}

According to this choice, the encoding of different parameters values is necessary for the suitable development of the selected technique. An appropriate coding of the chromosome is essential to achieve the optimal solution and this will depend on the kind of problem to solve. Our approach works with the first-level cache and both instruction and data cache can be customized with eligible values for each parameter.  Thus, a possible solution (individual) is defined as a specific cache configuration for I-cache and D-cache. Individual genes are then related to possible values of cache configuration parameters. Therefore, a chromosome is defined by the sequence of parameters for a specific cache configuration, coded as integer values. A chromosome, in our approach, looks like the one depicted in Figure \ref{fig:Chromosome}.

Thus, the chromosome applies an encoding scheme where each gene is an integer value that is mapped to the alphanumeric symbols (e.g. 8, 16, 32 for Line Size and LRU, FIFO, RANDOM for Replacement Policy) defined in Figure \ref{figure:arboles}, considering values from left to right, and starting with $0$.

\begin{figure*}
\begin{subfigure}
  \centering
  \includegraphics[width=0.95\textwidth]{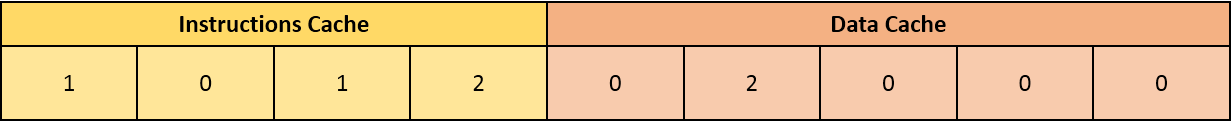}
  \caption{A chromosome or individual's genome.}
  \label{fig:ChromosomeExample} 
\end{subfigure}
\vspace{10mm}
\begin{subfigure}
  \centering
  \includegraphics[width=0.95\textwidth]{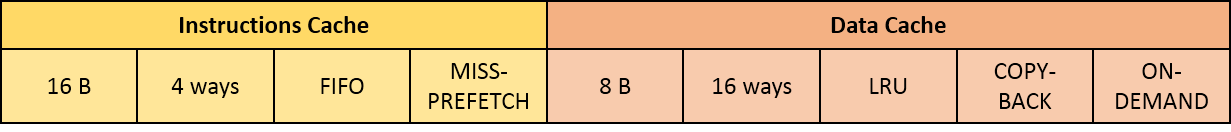}
  \caption{Decoded individual's genome.}
  \label{fig:ChromosomeDecoded} 
  \end{subfigure}
\end{figure*}

As an example, lets consider the chromosome in Figure~\ref{fig:ChromosomeExample}. The first and fifth genes ``1'' and ``0'' are the line sizes for the I-cache and D-cache, 16 and 8 bytes, respectively (following Figure \ref{figure:arboles}). Next, second and sixth genes ``0'' and ``2'' correspond to the degree of associativity, 4 and 16 ways. The third and seventh genes ``1'' and ``0'' correspond to the replacement algorithm, and are mapped to the FIFO and LRU algorithms. The fourth and eighth genes ``2'' and ``0'' are the MISS-PREFETCH and ON-DEMAND prefetch algorithms. Finally, the ninth gene ``0'' is mapped to copy-back policy. So, full genome decoded is shown in Figure~\ref{fig:ChromosomeDecoded}.

\section{Multi-objective function}\label{sec:multi}

Our approach defines a decision variable $\mathbf{x} \in \mathbf{X}$ in a multi-objective optimization context (see \ref{sec:AppA}). Variable $\mathbf{x}$ defines a set of cache parameters values that represent a cache configuration to evaluate. The evaluation process consists of calculating the multi-objective function $\mathbf{f}$ as the execution time and the energy consumption, both of them related to cache memory operations. Therefore, the best cache configuration will correspond with low values of execution time and energy consumption. As stated above, both are conflicting objectives. 

In order to evaluate different cache configurations we have applied energy and performance models based on~\cite{Janapsatya-2006}. So, the design of the embedded system architecture consists of a processor with one cache level with an instruction cache, a data cache and embedded DRAM as main memory. Both instruction and data caches are 16 Kbytes in size according to the default characteristics of the ARM920T processor~\cite{ARM}, our target platform. The instruction cache is read-only. Main memory is 64 MB in size according to datasheets of devices like Car GPS HS-3502, for example.  

\subsection{First objective: performance model}\label{sec:PerfModel}
The equation used to calculate execution time is described bellow. Execution time is computed according to time needed to solve accesses and misses on the cache memory system. 
\begin{eqnarray}
execTime & = & I_{access} \times I_{access\_time} +  \nonumber \\ 
  & & I_{miss} \times DRAM_{access\_time} + \nonumber \\
  & & I_{miss} \times I_{line\_size} \times \frac{1}{DRAM\_bw} + \nonumber \\
  & & D_{access} \times D_{access\_time} + \nonumber \\
  & & D_{miss} \times DRAM_{access\_time} + \nonumber \\
  & & D_{miss} \times D_{line\_size} \times \frac{1}{DRAM\_bw} \label{eq:extime}
\end{eqnarray}

\begin{itemize}
 \item $I_{access}$ and $D_{access}$ are the number of cache memory accesses to the instruction and data cache, respectively.
 \item $I_{miss}$ and $D_{miss}$ are the number of cache misses (when the data searched is not found in the cache memory and must be copied from the main memory).
 \item $I_{access\_time}$ and $D_{access\_time}$ represent the access time to the instruction and data cache respectively per access.
 \item $DRAM_{access\_time}$ is the main memory latency time.
 \item $I_{line\_size}$ and $D_{line\_size}$ correspond to line size (or block size) for instruction a data cache, respectively. 
 \item $DRAM\_{bw}$ is the bandwidth of the DRAM (transfer capacity).
\end{itemize}

This equation have six well defined parts, as detailed in~\cite{Janapsatya-2006}. Therefore, $I_{access} \times I_{access\_time}$ is the total time due to instruction cache accesses.
$I_{miss} \times DRAM_{access\_time}$ is the total time spent by main memory accesses in response to instruction cache misses. 
$I_{miss} \times I_{line\_size} \times \frac{1}{DRAM\_bw}$ represents the total time needed to fill a cache line for each cache miss on the instructions cache.
$D_{access} \times D_{access\_time}$ is the total time due to data cache accesses. 
$D_{miss} \times DRAM_{access\_time}$ is the total time spent by main memory accesses in response to data cache misses and 
$D_{miss} \times D_{line\_size} \times \frac{1}{DRAM\_bw}$ represents the total time needed to fill a cache line for each cache miss on the data cache.

\subsection{Second objective: energy model}
Energy model is explained according to the following equation: 
\begin{eqnarray}
Energy & = & execTime \times CPU_{power} + \nonumber \\ 
& & I_{access} \times I_{access\_energy} +  \nonumber \\ 
& & D_{access} \times D_{access\_energy} + \nonumber \\
& & I_{miss} \times I_{access\_energy} \times I_{line\_size} + \nonumber \\
& & D_{miss} \times D_{access\_energy} \times D_{line\_size} + \nonumber \\
& & I_{miss} \times DRAM_{access\_power} \times \nonumber \\
& & (\small{DRAM_{access\_time}} + \small{I_{line\_size}} \times \frac{1}{\small{DRAM_{bw}}}) + \nonumber \\
& & D_{miss} \times DRAM_{access\_power} \times \nonumber \\
& & (\small{DRAM_{access\_time}} + \small{D_{line\_size}} \times \frac{1}{\small{DRAM_{bw}}}) \label{eq:energy1}
\end{eqnarray}
where varibles not described in Section \ref{sec:PerfModel} are:
\begin{itemize}
 \item $DRAM_{access\_power}$ is the power consumption for each DRAM access.
 \item $I_{access\_energy}$  and $D_{access\_energy}$ correspond to energy consumption in each instruction and data cache access, respectively.
\end{itemize}

The $I_{access} \times I_{access\_energy}$ and $D_{access} \times D_{access\_energy}$ terms calculate the energy consumption because of instructions and data cache, respectively. $I_{miss} \times I_{access\_energy} \times I_{line\_size}$ and $D_{miss} \times D_{access\_energy} \times D_{line\_size}$ is the energy cost of filling information into instruction and data caches respectively from main memory when miss occurs. The last two terms calculate the energy cost of the DRAM to respond to cache misses.

In our approach we remove the first term of the Energy equation $execTime \times CPU_{power}$ because of three reasons: (1) the term $CPU_{power}$ is constant and the term $execTime$ is already being minimized in the first objective, (2) it represents the amount of energy consumed by the CPU and we are optimizing just the performance and energy consumed by the memory subsystem, and (3) in a multi-objective optimization all the objectives must be as orthogonal as possible, i.e., the term $execTime$ is redundant. Thus, our second objective is reduced to:

\begin{eqnarray}
Energy & = & I_{access} \times I_{access\_energy} +  \nonumber \\ 
& & D_{access} \times D_{access\_energy} + \nonumber \\
& & I_{miss} \times I_{access\_energy} \times I_{line\_size} + \nonumber \\
& & D_{miss} \times D_{access\_energy} \times D_{line\_size} + \nonumber \\
& & I_{miss} \times DRAM_{access\_power} \times \nonumber \\
& & (\small{DRAM_{access\_time}} + \small{I_{line\_size}} \times \frac{1}{\small{DRAM_{bw}}}) + \nonumber \\
& & D_{miss} \times DRAM_{access\_power} \times \nonumber \\
& & (\small{DRAM_{access\_time}} + \small{D_{line\_size}} \times \frac{1}{\small{DRAM_{bw}}}) \label{eq:energy}
\end{eqnarray}

All the equations use seconds for time, watts for power, Joules for energy, bytes for cache line size and bytes/sec for bandwidth.

Our algorithm evolves to minimize execution time and/or energy consumption. After a given number of generations the algorithm returns a Pareto Front (an approximation to the Pareto Optimal Front), that represents the best set of configurations to apply to the cache memory. The higher the number of generations, the better is the quality of the cache memory.

\section{Optimization framework}\label{sec:Methodology}

\begin{figure*}[ht]
  \centering
  \includegraphics[width=0.98\textwidth]{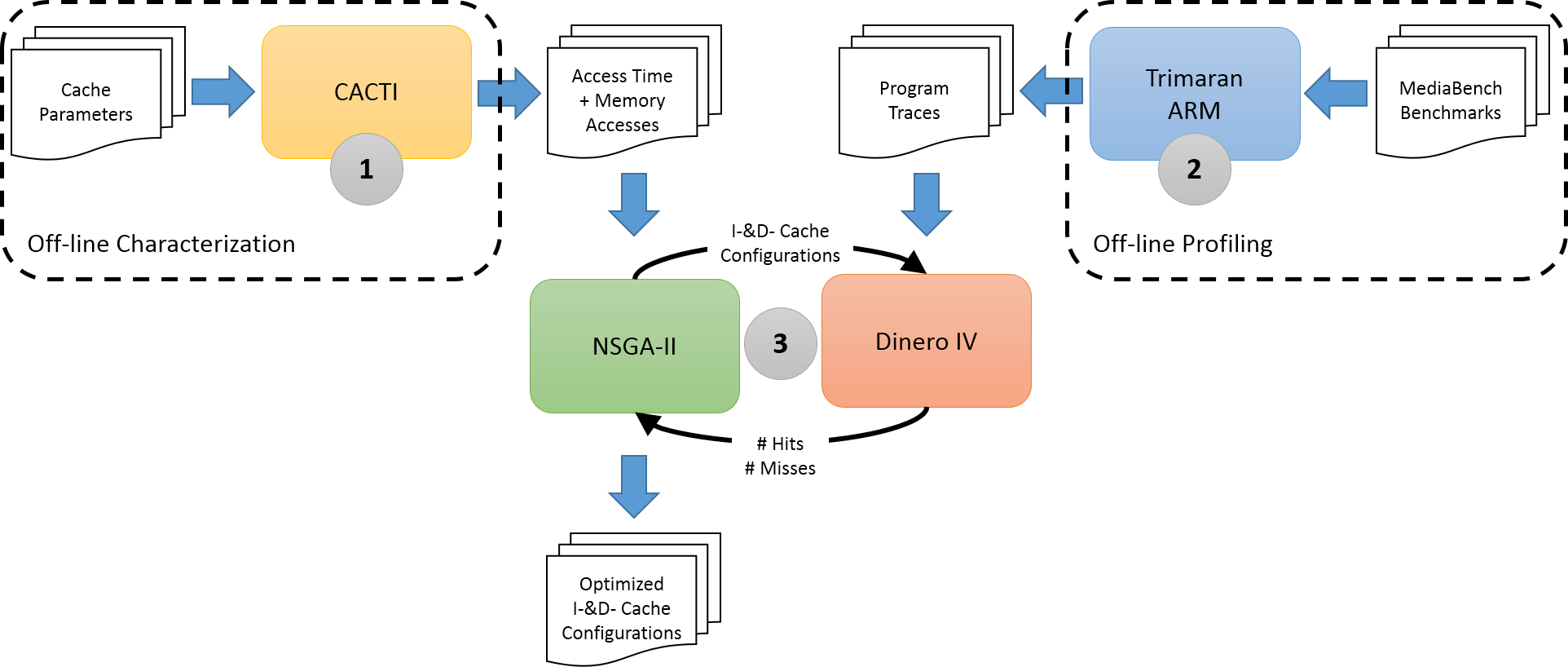}
  \caption{Three processes are involved in the cache configuration optimization: (1) cache characterization, (2) application profiling and (3) cache optimization.} 
  \label{figure:method}
\end{figure*}

In this section we describe the framework used to optimize cache memories for multimedia embedded systems. As mentioned above, this work proposes an approach to determine the best cache configurations for a given set of applications. The best cache configurations are those which take less execution time and less energy consumption. Figure \ref{figure:method} depicts all the steps needed to carry out the optimization process. 

We have divided our optimization process into three different phases, labeled in Figure \ref{figure:method}. Firstly, two processes are executed just once before the optimization (labeled as 1 and 2). Next, the optimization is performed (labeled as 3), using as input the results of the previous two phases. We have extracted the first two off-line phases from the optimization phase to save execution time. The first phase is performed in one hour, whereas the second phase can be completed in four hours. Using this pre-characterization policy saves months to the optimization process (more details of execution time are provided in Section \ref{sec:Experiments}). In the following, we describe more in depth these three phases.

The first phase is \emph{cache characterization}. The characterization of the DRAM and cache memory is performed using Cacti~\cite{Mamidipaka2004} to compute access times and energy. Cacti is a widely used analytical model to estimate energy and power consumption, performance and area of caches. The characterization is performed off-line. Basic inputs required by Cacti are cache size, line size and associativity. Since the cache size is fixed, and line size and degree of associativity have 4 possible values, the number of possible cache characterizations is 16. After this phase, all the parameters needed in the objective function (equations \ref{eq:extime} and \ref{eq:energy}) are available. 

The second phase is \emph{application profiling}. All the target applications are simulated with Trimaran and all cache memory accesses are compiled and saved in program traces. Trimaran is an integrated compilation and performance monitoring infrastructure which provides enough resources to obtain application traces with accuracy. Trimaran customizes ARM processors through simpleScalar~\cite{simplescalar}, an architectural simulator that can model a large set of different architectures. The processing time required to perform this phase depends on the number of target applications and the number of instructions to simulate.

The third phase is \emph{cache optimization}. This phase must be repeated for each target application and is carried out by the NSGA-II algorithm implemented in the JECO library~\cite{JECO}. NSGA-II evaluates every candidate solution calling Dinero IV, which is a trace-driven cache simulator~\cite{DineroIV}. Dinero IV receives a cache configuration from NSGA-II and returns the number of cache hits and misses for the corresponding target application trace. These data, and the parameters obtained in the first phase are then included in the multi-objective function to compute both the execution time and energy consumed.

We have selected NSGA-II as the multi-objective optimization algorithm because, according to a recent survey published in \cite{Sayyad2013}, the current de facto standard evolutionary algorithm for multi-objective optimization is NSGA-II. This survey states that NSGA-II was used as a single algorithm in 53\% of the examined papers, positioning the algorithm as one of the most widely used MOEAs, and obtaining very competitive results. Since our aim is to provide a technique to automatically design optimized cache memories, and not to find the best optimization algorithm, we propose the use of this one.

\begin{figure*}[ht]
  \centering
  \includegraphics[width=0.65\textwidth]{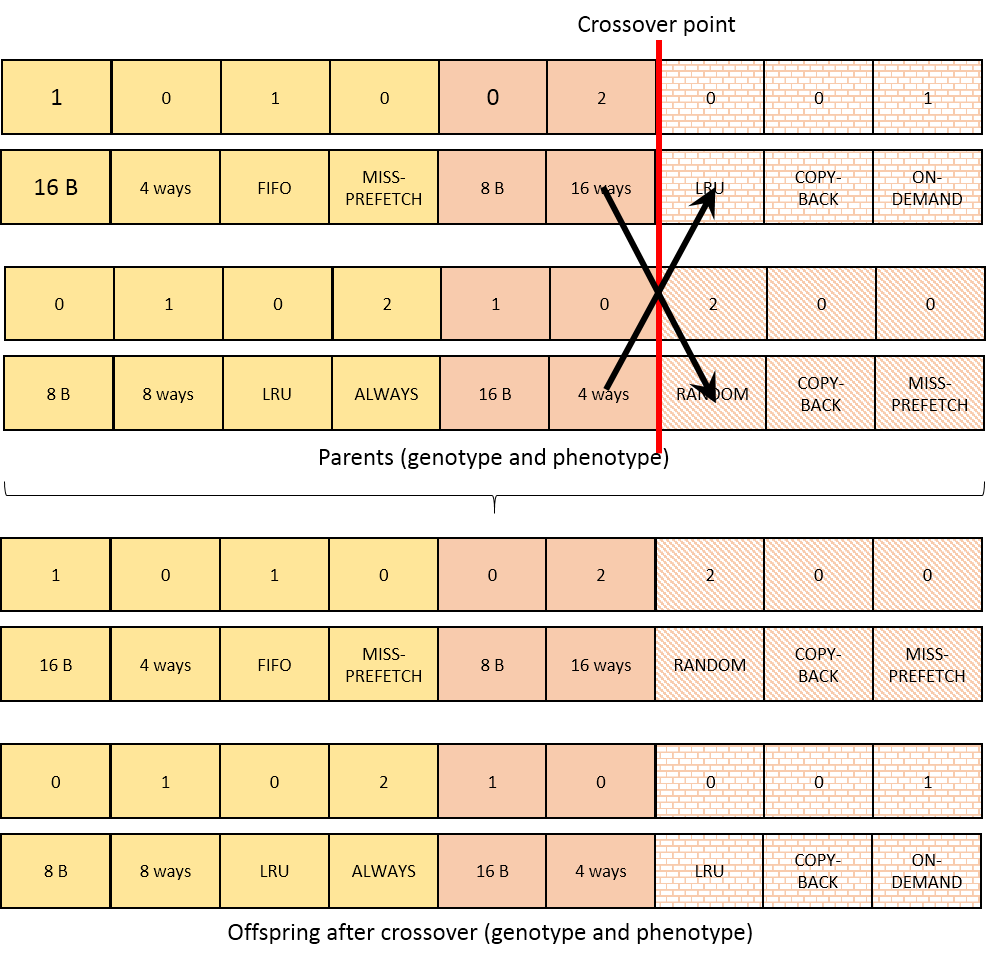}
  \caption{Single point crossover.} 
  \label{fig:Crossover}
\end{figure*}

\begin{figure*}[ht]
  \centering
  \includegraphics[width=0.65\textwidth]{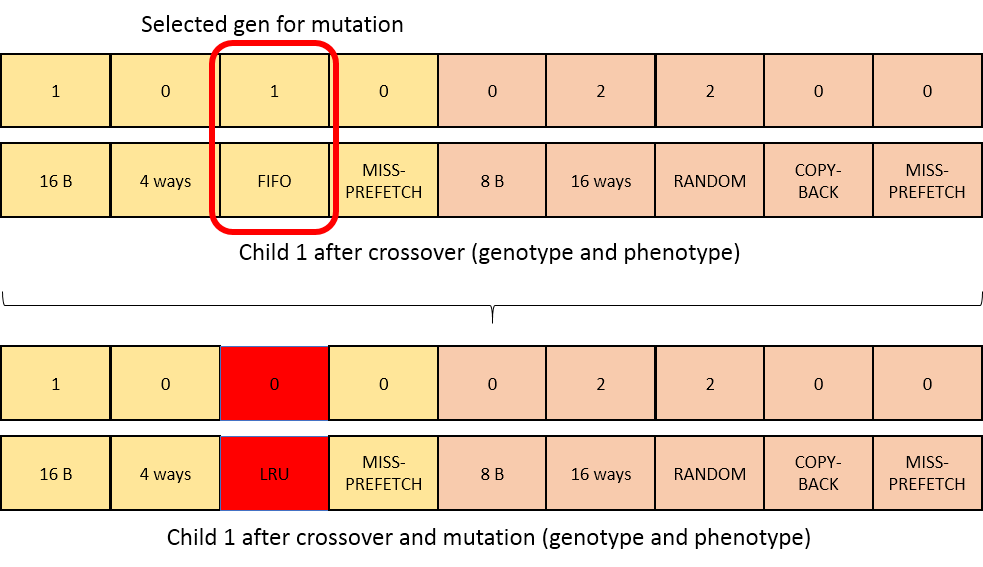}
  \caption{Integer flip mutation.} 
  \label{fig:Mutation}
\end{figure*}

For NSGA-II, we have used single point crossover and integer flip mutation operators. The single point crossover is illustrated in Figure \ref{fig:Crossover}, where a random point is selected in the chromosome and used to generate two children. Similarly, the integer flip mutation is depicted in Figure \ref{fig:Mutation}. A random integer is generated for all those genes that must be mutated (according to the mutation probability), always constrained to the limits of the corresponding gene. Following the example given in Figure \ref{fig:Mutation}, the third gene is mutated, modifying its value from ``1'' to ``0'', which in the phenotype is translated into a change from FIFO to LRU replacement algorithm, respectively.

\section{Experiments}\label{sec:Experiments}

Our simulation environment consists of an Intel(R) Core(TM) i7-3770 CPU @ 3.40GHz with 16 GB RAM memory, with a GNU/Linux Debian 7 Operating System running a master-worker parallel version of NSGA-II with 8 workers. Experimental results are based on the ARM architecture. ARM processors are widespread on multimedia embedded devices. ARM920T~\cite{ARM} is a typical embedded processor used in tablets, smartphones, and set-top boxes like Motorola Q9m Verizon Mobile Phone, Car GPS HS-3502, etc. The ARM920T processor is a member of the ARM9TDMI family of general-purpose  microprocessors, which have a standalone processor core on a Harvard architecture device. By default, the ARM920T processor implements a separate 16 KB instruction and data cache.

\subsection{Setup}
To evaluate the effectiveness of our approach we have selected a subset of the Mediabench~\cite{mediabench} applications suite as our target applications. Although our methodology can be used with any kind of applications, the Mediabench benchmark has been selected because of the high variability in block size, which provides heterogeneity to the exploration space. Using our methodology, we design an optimal first level cache memory with fixed-size for instructions and data, similar to some devices that have an ARM920T processor. After that, we have validated our optimization framework with two additional hardware platforms. We have simulated twelve Mediabench benchmarks: cjpeg, djpeg, mpegdec, mpegenc, gsmdec, gsmenc, epic, unepic, pegwitdec, pegwitenc, rawcaudio and rawdaudio, all of them with their standard input. As stated above, we have generated their traces using Trimaran tools~\cite{trimaran}. Trimaran works with SimpleScalar~\cite{simplescalar}. Thus, we have modified both SimpleScalar and Trimaran tools to obtain application traces according to the Dinero IV cache simulator, which is continuously called by our parallel NSGA-II implementation to evaluate each candidate solution. 

Every application has been simulated for $7.5 \times 10^7$ instructions to reach a balance between the simulation time, the size of the program traces generated and a proper number of instructions. NSGA-II has been executed 30 times for each target application. 

\begin{table}
\centering
\begin{tabular}{|c|c|}
\hline
Number of generations & $250$ \\
\hline
Population size & $100$ \\ 
\hline
Chromosome length & $9$ \\ 
\hline
Probability of crossover & $0.9$ \\ 
\hline
Probability of mutation & $1/9$ \\ 
\hline
\end{tabular}
\caption{NSGA-II Algorithm Parameters.}
\label{Table:genetic}
\end{table}

Table~\ref{Table:genetic} shows the NSGA-II configuration. As crossover and mutation probabilities, we have used the values recommended in \cite{Deb2002}. The number of generations and individuals has been fixed after several tests.

\subsection{Optimization Results}

\begin{figure*}[ht]
  \includegraphics[width=0.99\textwidth]{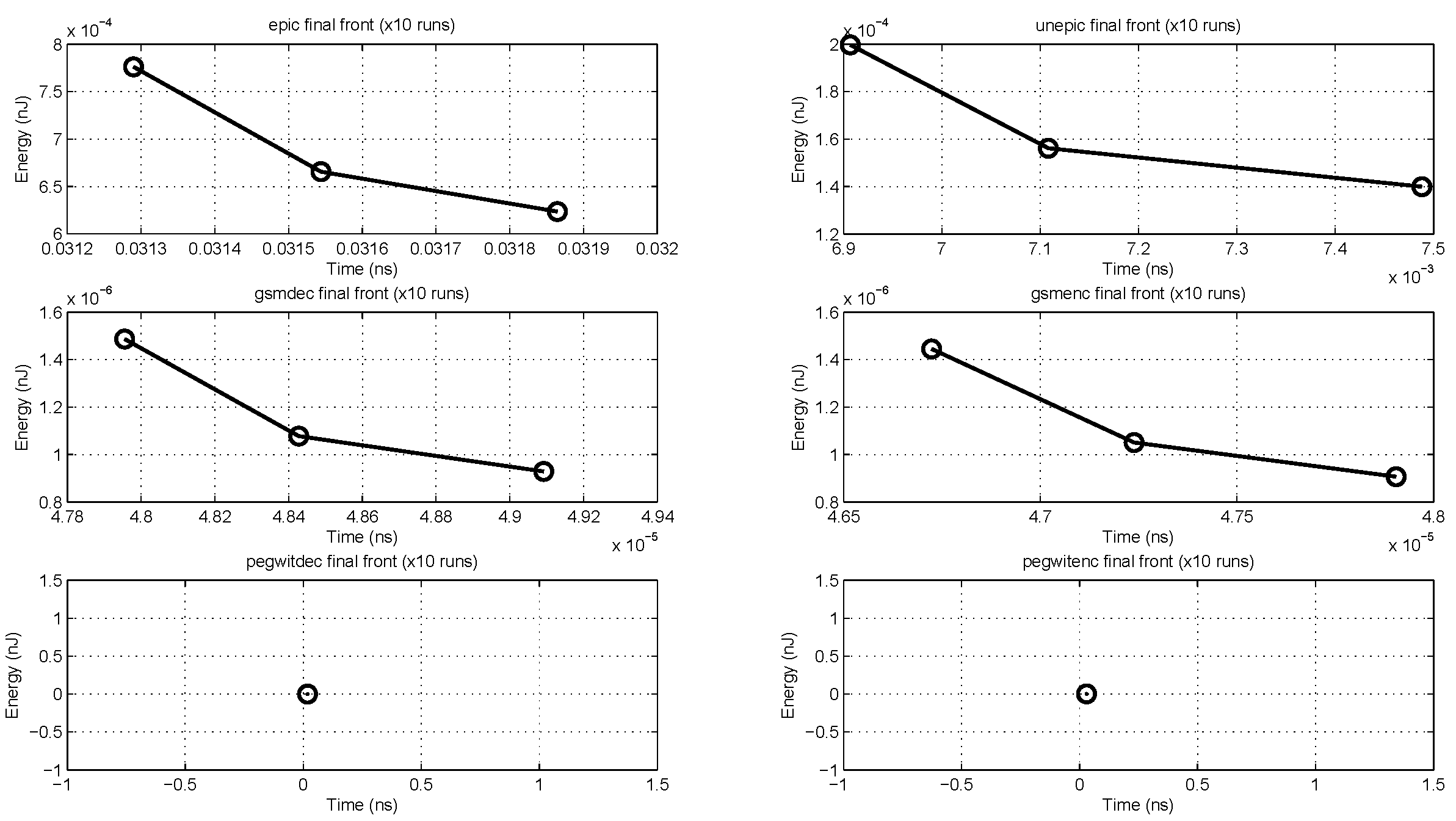}
  \caption{Pareto front representation for epic, unepic, gsmdec, gsmenc, pegwitdec and pegwitenc.}
  \label{Figure1}
\end{figure*}

\begin{figure*}[ht]
  \centering
  \includegraphics[width=0.99\textwidth]{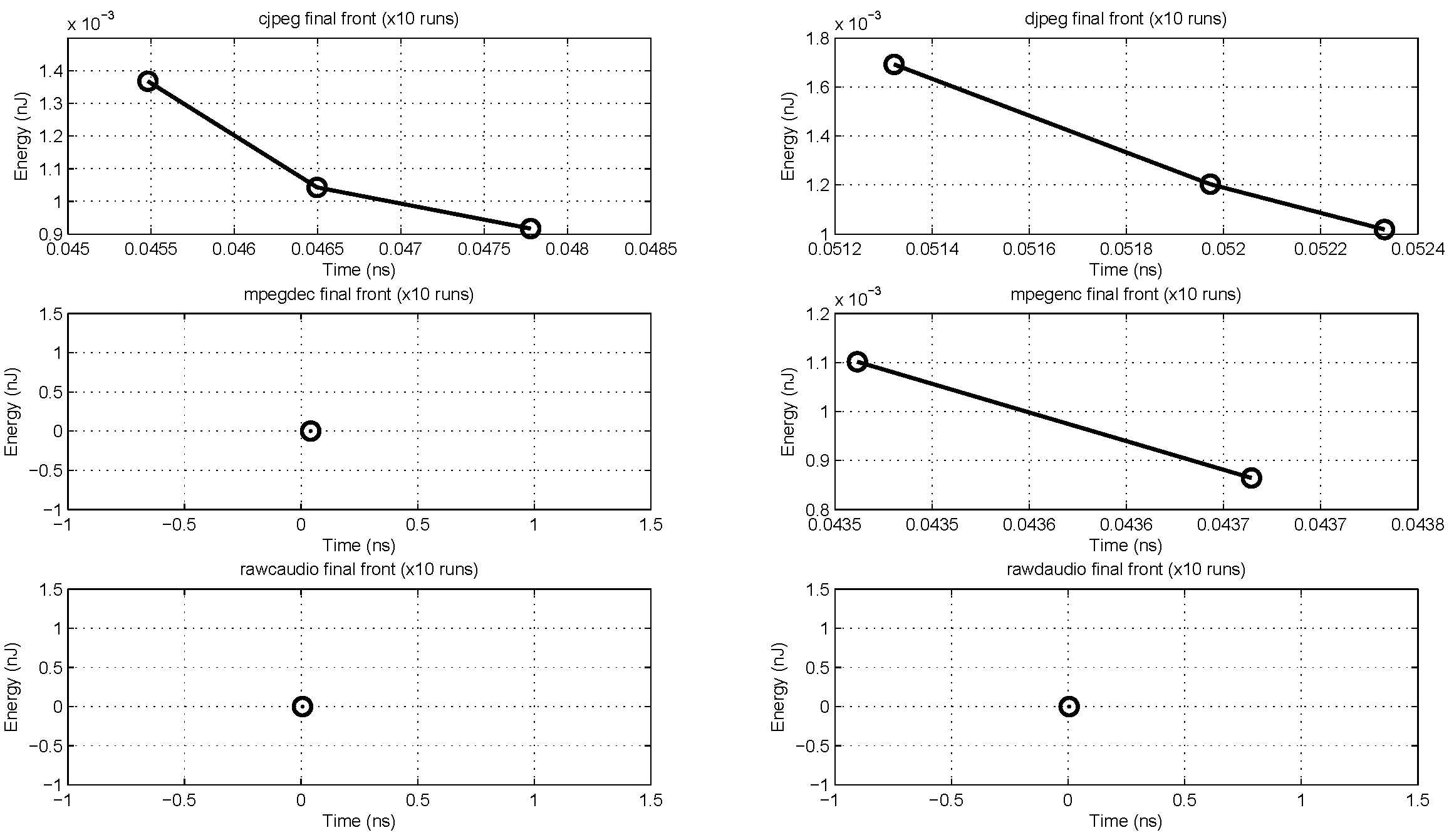}
  \caption{Pareto front representation for cjpeg, djpeg, mpegdec, mpegenc, rawcaudio and rawdaudio.}
  \label{Figure2}
\end{figure*}

In the following we show and analyze all the results obtained in this research work. Figures~\ref{Figure1} and \ref{Figure2} show the Pareto fronts obtained with our optimization framework. Each point in the graph represents a cache configuration and the corresponding execution time and energy, driven by equations (\ref{eq:extime}) and (\ref{eq:energy}), respectively. 

From Figures~\ref{Figure1} and \ref{Figure2}, we may observe that the algorithm obtains at least one optimized cache configuration for each application. Indeed we confirm our hypothesis that both execution time and energy are conflicting objectives. It is worth noting that we can see a single point in some plots, which seems to represent a single solution instead of a Pareto front. However, these single points represent more than one cache configuration with the same values in the objective space. To illustrate this point, we show in Table~\ref{paretos} the Pareto set obtained for each application, along with their respective objective values. The mpegdec application, for example, shows two different cache configurations for the same objective values. The same happens with rawcaudio, with 17 different configurations with same objective values. This is an excellent result, because the selection of a good cache configuration for all the target applications is simplified.

\begin{table*}
\centering
\tiny
\begin{tabular}{|c|c|c|c|c|c|c|c|c|c|c|c|}
\hline
\textbf{Application} & \textbf{LI} & \textbf{WI} & \textbf{RI} & \textbf{SI} & \textbf{LD} & \textbf{WD} & \textbf{RD} &\textbf{AD}&\textbf{SD} & \textbf{ExTime} & \textbf{Energy} 
\\ \hline
\multirow{2}{*}{mpegdec}   & \multirow{2}{*}{8}  & \multirow{2}{*}{4}  & \multirow{2}{*}{RANDOM} & \multirow{2}{*}{Always} & \multirow{2}{*}{8}  & \multirow{2}{*}{4}  & \multirow{2}{*}{LRU}    & Write-back    & \multirow{2}{*}{Miss-prefetch} & 0.04116      & 0.00082 \\ 
 & & & & & & & & Write-through & & 0.04116 & 0.00082 \\ 
\hline
\multirow{4}{*}{mpegenc} & \multirow{4}{*}{8} & \multirow{4}{*}{4} & \multirow{4}{*}{LRU} & \multirow{4}{*}{Always} & \multirow{2}{*}{8} & \multirow{4}{*}{4}  & \multirow{4}{*}{LRU} & Write-back & \multirow{4}{*}{Miss-prefetch} & 0.04371      & 0.000864 \\
          & & & & & & & & Write-through & & 0.04371      & 0.000864 \\
					\cline{6-6}
          & & & & & \multirow{2}{*}{32} & & & Write-back & & 0.04351      & 0.001101 \\
          & & & & & & & & Write-through & & 0.04351      & 0.001101 \\ 
\hline
\multirow{6}{*}{jpeg} & \multirow{6}{*}{8} & \multirow{6}{*}{4} & \multirow{6}{*}{LRU} & \multirow{6}{*}{Always} & \multirow{2}{*}{8} & \multirow{6}{*}{4} & \multirow{6}{*}{LRU} & Write-back & \multirow{6}{*}{Miss-prefetch} & 0.04778      & 0.000917 \\
          & & & & & & & & Write-through & & 0.04778      & 0.000917 \\
					\cline{6-6}
    	    & & & & & \multirow{2}{*}{16} & & & Write-back & & 0.04649      & 0.001043 \\
    	    & & & & & & & & Write-through & & 0.04649      & 0.001043 \\
					\cline{6-6}
    	    & & & & & \multirow{2}{*}{32} & & & Write-back & & 0.04548      & 0.001367 \\
    	    & & & & & & & & Write-through & & 0.04548      & 0.001367 \\ 
\hline
\multirow{6}{*}{djpeg} & \multirow{6}{*}{8} & \multirow{6}{*}{4}   & \multirow{6}{*}{LRU}    & \multirow{6}{*}{Always} & \multirow{2}{*}{8}  & \multirow{6}{*}{4}  & \multirow{6}{*}{LRU}    & Write-back    & \multirow{6}{*}{Miss-prefetch} & 0.05233      & 0.00102 \\
    	    & 	&  &  &  &  &  &  & Write-through &  & 0.05233 &	0.00102\\
					\cline{6-6}
          &  &  &  &  & \multirow{2}{*}{16} &  &  & Write-back &  & 0.05197 & 0.00120\\
          &  &  &  &  &  &  &  & Write-through &  & 0.05197 & 0.00120 \\
					\cline{6-6}
          &  &  &  &  & \multirow{2}{*}{32} &  &  & Write-back &  & 0.05132 & 0.00169 \\
          &  &  &  &  &  &  &  & Write-through &  & 0.05132 & 0.00169\\ 
\hline
\multirow{6}{*}{epic}      & \multirow{6}{*}{8}	& \multirow{6}{*}{4} & \multirow{6}{*}{LRU} & \multirow{6}{*}{Always} & \multirow{2}{*}{8} & \multirow{6}{*}{4} & \multirow{6}{*}{LRU} & Write-back & \multirow{6}{*}{Miss-prefetch} & 0.03187 & 0.00062 \\
          &  &  &  &  &  &  &  & Write-through &  & 0.03187 & 0.00062 \\
					\cline{6-6}
          &  &  &  &  & \multirow{2}{*}{16} &  & & Write-back &  & 0.03154 & 0.00067 \\
          &  &  &  &  &  &  & & Write-through &  & 0.03154 & 0.00067 \\
					\cline{6-6}
          &  &  &  &  & \multirow{2}{*}{32} &  & & Write-back &  & 0.03129 & 0.00078 \\
          &  &  &  &  &  &  & & Write-through &  & 0.03129 & 0.00078 \\ 
\hline
\multirow{8}{*}{unepic}    & \multirow{8}{*}{8}  & \multirow{8}{*}{4}  & \multirow{6}{*}{LRU}    & \multirow{8}{*}{Always} & \multirow{2}{*}{8}  & \multirow{8}{*}{4}  & \multirow{2}{*}{LRU}    & Write-back & \multirow{8}{*}{Miss-prefetch} & 0.007488 & 0.0001399 \\
          &   &   &     &  &   &   &     & Write-through & & 0.007488 & 0.0001399 \\
					\cline{6-6}\cline{8-8}
    	    &   &   &     &  & \multirow{2}{*}{16} &  & \multirow{6}{*}{RANDOM} & Write-back    &  & 0.007108 & 0.0001561 \\
    	    &   &   &     &  &  &   &  & Write-through &  & 0.007108 & 0.0001561 \\
					\cline{6-6}
    	    &   &   &     &  & \multirow{4}{*}{32} &   &  & Write-back &  & 0.0069067 & 0.0001997 \\
    	    &   &   &     &  &  &   &  & Write-through    &  & 0.0069067 & 0.0001997 \\
					\cline{4-4}
    	    &   &   & \multirow{2}{*}{RANDOM} &  &  &   &  & Write-back &  & 0.0069068 & 0.0001997 \\
    	    &   &   &  &  &  &   &  & Write-through &  & 0.0069068 & 0.0001997 \\ 
\hline
\multirow{16}{*}{rawcaudio} & \multirow{16}{*}{8} & \multirow{16}{*}{4} & \multirow{6}{*}{FIFO} & \multirow{16}{*}{Always} & \multirow{16}{*}{8} & \multirow{16}{*}{4} & \multirow{2}{*}{FIFO} & Write-back & \multirow{16}{*}{Miss-prefetch} & 0.00673 & 0.00013 \\
          &  &  &  &  &  &  &  & Write-through & & 0.00673 & 0.00013 \\
					\cline{8-8}
          &  &  &  &  &  &  & \multirow{2}{*}{LRU} & Write-back & & 0.00673 & 0.00013  \\
          &  &  &  &  &  &  &  & Write-through & & 0.00673 & 0.00013 \\
					\cline{8-8}
          &  &  &  &  &  &  & \multirow{2}{*}{RANDOM} & Write-back & & 0.00673 & 0.00013 \\
          &  &  &  &  &  &  &  & Write-through & & 0.00673 & 0.00013 \\
					\cline{4-4}\cline{8-8}
          &  &  & \multirow{6}{*}{LRU} &  &  &  & \multirow{2}{*}{FIFO} & Write-back & & 0.00673 & 0.00013 \\
          &  &  &  &  &  &  &  & Write-through & & 0.00673 & 0.00013 \\
					\cline{8-8}
          &  &  &  &  &  &  & \multirow{2}{*}{LRU} & Write-back & & 0.00673 & 0.00013 \\
          &  &  &  &  &  &  &  & Write-through & & 0.00673 & 0.00013 \\
					\cline{8-8}
          &  &  &  &  &  &  & \multirow{2}{*}{RANDOM} & Write-back & & 0.00673 & 0.00013 \\
          &  &  &  &  &  &  &  & Write-through & & 0.00673 & 0.00013 \\
					\cline{4-4}\cline{8-8}
          &  &  & \multirow{4}{*}{RANDOM} &  &  &  & \multirow{2}{*}{FIFO} & Write-back & & 0.00673 & 0.00013 \\
          &  &  &  &  &  &  &  & Write-through & & 0.00673 & 0.00013 \\
					\cline{8-8}
          &  &  &  &  &  &  & \multirow{2}{*}{LRU} & Write-back & & 0.00673 & 0.00013 \\
          &  &  &  &  &  &  &  & Write-through & & 0.00673 & 0.00013 \\
\hline
\multirow{8}{*}{rawdaudio} & \multirow{8}{*}{8} & \multirow{8}{*}{4} & \multirow{6}{*}{LRU} & \multirow{8}{*}{Always} & \multirow{8}{*}{8} & \multirow{8}{*}{4} & \multirow{2}{*}{FIFO} & Write-back & \multirow{8}{*}{Miss-prefetch} & 0.00492 & 0.000098 \\
          &  &  &  &  &  &  &  & Write-through & & 0.00492 & 0.000098 \\
					\cline{8-8}
          &  &  &  &  &  &  & \multirow{2}{*}{LRU} & Write-back & & 0.00492 & 0.000098 \\
          &  &  &  &  &  &  &  & Write-through & & 0.00492 & 0.000098 \\
					\cline{8-8}
          &  &  &  &  &  &  & \multirow{4}{*}{RANDOM} & Write-back & & 0.00492 & 0.000098 \\
          &  &  &  &  &  &  &  & Write-through & & 0.00492 & 0.000098 \\
					\cline{4-4}
          &  &  & \multirow{2}{*}{RANDOM} &  &  &  &  & Write-back & & 0.00492 & 0.000098 \\
          &  &  &  &  &  &  &  & Write-through & & 0.00492 & 0.000098 \\ 
\hline
\multirow{6}{*}{gsmdec}    & \multirow{6}{*}{8} & \multirow{6}{*}{4} & \multirow{6}{*}{LRU} & \multirow{6}{*}{Always} & \multirow{2}{*}{8} & \multirow{6}{*}{4} & \multirow{6}{*}{LRU} & Write-back & \multirow{6}{*}{Miss-prefetch} & 4.91E-005 & 9.29E-007 \\
          &  &  &  &  &  &  &  & Write-through &  & 4.91E-005 & 9.29E-007 \\
					\cline{6-6}
          &  &  &  &  & \multirow{2}{*}{16} &  &  & Write-back &  & 4.84E-005 & 1.08E-006 \\
          &  &  &  &  &  &  &  & Write-through &  & 4.84E-005 & 1.08E-006 \\
					\cline{6-6}
          &  &  &  &  & \multirow{2}{*}{32} &  &  & Write-back &  & 4.80E-005 & 1.49E-006 \\
          &  &  &  &  &  &  &  & Write-through &  & 4.80E-005 & 1.49E-006 \\ 
\hline
\multirow{6}{*}{gsmenc}    & \multirow{6}{*}{8} & \multirow{6}{*}{4} & \multirow{6}{*}{LRU} & \multirow{6}{*}{Always} & \multirow{2}{*}{8} & \multirow{6}{*}{4} & \multirow{6}{*}{LRU} & Write-back & \multirow{6}{*}{Miss-prefetch} & 0.000048 & 0.0000009 \\ 
          &  &  &  &  &  &  &  & Write-through &  & 0.000048 & 0.0000009 \\
					\cline{6-6}
          &  &  &  &  & \multirow{2}{*}{16} &  &  & Write-back &  & 0.000047 & 0.0000011 \\
          &  &  &  &  &  &  &  & Write-through &  & 0.000047 & 0.0000011 \\
					\cline{6-6}
          &  &  &  &  & \multirow{2}{*}{32} &  &  & Write-back &  & 0.000047 & 0.0000014 \\
          &  &  &  &  &  &  &  & Write-through &  & 0.000047 & 0.0000014 \\ 
\hline
\multirow{2}{*}{pegwitdec} & \multirow{2}{*}{8} & \multirow{2}{*}{4} & \multirow{2}{*}{RANDOM} & \multirow{2}{*}{Always} & \multirow{2}{*}{8} & \multirow{2}{*}{4} & \multirow{2}{*}{LRU} & Write-back & \multirow{2}{*}{On-demand} & 0.01795 & 0.00034 \\
          &  &  &  &  &  &  &  & Write-through &  & 0.01795 & 0.00034 \\ 
\hline
\multirow{2}{*}{pegwitenc} & \multirow{2}{*}{8} & \multirow{2}{*}{4} & \multirow{2}{*}{LRU} & \multirow{2}{*}{Always} & \multirow{2}{*}{8} & \multirow{2}{*}{4} & \multirow{2}{*}{LRU} & Write-back & \multirow{2}{*}{On-demand} & 0.02952 & 0.00055 \\
          &  &  &  &  &  &  &  & Write-through &  & 0.02952 & 0.00055 \\ 
\hline
\end{tabular}
\caption{Pareto Sets for a Cache Size of 16 KB.}
\label{paretos}
\end{table*}

\begin{figure*}
 \centering
  \includegraphics[width=0.95\textwidth]{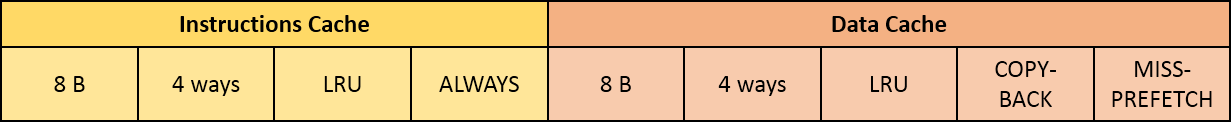}
  \caption{Cache Configuration shared by all the Pareto sets.}
  \label{fig:ParetoSets}
\end{figure*}

In this regard, Table~\ref{paretos} shows that there are two cache configurations found in nine out of the twelve applications under study. One of these two configurations is shown in Figure \ref{fig:ParetoSets}.

Mpegdec, pegwitdec and pegwitenc are the only applications that do not share this cache configuration. The best configurations found save a $63.45\%$  and $91.68\%$ for mpegdec, $60.09\%$ and $92.46\%$ for pegwitdec and $59.74\%$ and $92.55\%$ for pegwitenc in execution time and energy, respectively. However, we have detected that using the cache configuration of Figure \ref{fig:ParetoSets}, we save almost the same quantities in execution time and energy ($61.39\%$ and $90.98\%$ for mpegdec, $60.05\%$ and $91.83\%$ for pegwitdec and $58.89\%$ and $92.14\%$ for pegwitenc). Definitely, these are really good results to unify the selection process of an optimized cache configuration, for a target set of applications.

However, the number or points in the final Pareto front is small compared to the size of the search space (more than 64000 alternatives, as computed in Section \ref{sec:dspace}). It might occur because NSGA-II have found the global optimum or the algorithm usually falls into a strong local optimum. To clarify this point, we have computed the hypervolume indicator ($I_H^-$) for each single run (see \ref{sec:AppA}). 

Table~\ref{table_h} shows all the hypervolumes averaged for most of the 30 different runs and 7 applications. We did not compute the hypervolume indicator for all the target applications because in some of them we just obtained the same single solution in each of the 30 runs, and the hypervolume cannot be computed for one single point. In the remaining cases, it is worth noting that the standard deviation is almost 0 for all the seven applications, i.e., NSGA-II is finding the same Pareto front on each simulation. Given that the algorithm always started with a different random initial population, it probably means that NSGA-II reached the \emph{Pareto-Optimal} front.

\begin{table}
\centering
\begin{tabular}{|ccc|}
\hline
\ \textbf{Application} & \textbf{Mean} & \textbf{STD}\\ 
\hline
epic       & $-0.61$  & 0 \\ 
unepic     & $-0.69$  & 0 \\ 
jcpeg      & $-0.61$  & 0 \\ 
jdpeg      & $-0.47$  & 0 \\ 
gsmdec     & $-0.64$  & 0 \\ 
gsmenc     & $-0.62$  & $1.18 \times 10^{-16}$ \\ 
mpegenc    & $-0.21$  & 0 \\ 
\hline
\end{tabular}
\caption{Hypervolume Metric (S-metric).}
\label{table_h}
\end{table}

\subsection{Comparison with a baseline cache}
To analyze the level of improvement using our optimization framework, we compare our results with those obtained by a baseline cache configuration. The selected baseline configuration appears in devices mentioned above (Motorola Q9m Verizon Mobile Phone, Car GPS HS-3502, among others). This cache has the following configuration values:

\begin{itemize}
 \item Icache: Cache size: 16 KB; Block size: 16; Associativity: 64; Replacement algorithm: LRU: Prefetch policy: ON-DEMAND; 
 \item Dcache: Cache size: 16 KB; Block size: 16; Associativity: 64; Replacement algorithm: LRU: Prefetch policy: ON-DEMAND; Write policy: COPY-BACK;
\end{itemize}

We have computed execution time and energy for this baseline cache following our model developed in Section \ref{sec:multi}. Next, we compare each point in the Pareto fronts depicted in Figures \ref{Figure1} and \ref{Figure2} with the baseline metrics using the following equations:

\begin{eqnarray}
\mathrm{Improvement}_{\mathrm{execTime}} & = & 100 \times \frac{T_{\mathrm{baseline}}-T_{\mathrm{optimized}}}{T_{\mathrm{baseline}}} \\
\mathrm{Improvement}_{\mathrm{Energy}} & = & 100 \times \frac{E_{\mathrm{baseline}}-E_{\mathrm{optimized}}}{E_{\mathrm{baseline}}}
\end{eqnarray}
where $T_{\mathrm{baseline}}$ and $T_{\mathrm{optimized}}$ are the execution time of the baseline and optimized caches, respectively. In the same manner, $E_{\mathrm{baseline}}$ and $E_{\mathrm{optimized}}$ are the energy computed for baseline and optimized caches, respectively.

\begin{figure*}[ht]
  \centering
  \includegraphics[width=0.95\textwidth]{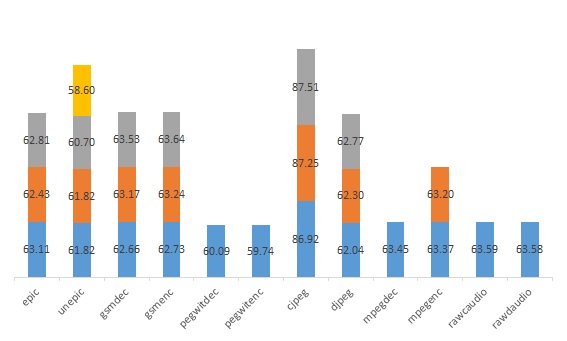}
  \caption{Pareto front execution time with respect to the baseline configuration (labels represent the percentage of improvement in execution time). Each color represents one point in the non-dominated front. For example, unepic has 4 points, whereas pegwitdec has one single point.}
  \label{Figure3}
\end{figure*}

\begin{figure*}[ht]
  \centering
  \includegraphics[width=0.95\textwidth]{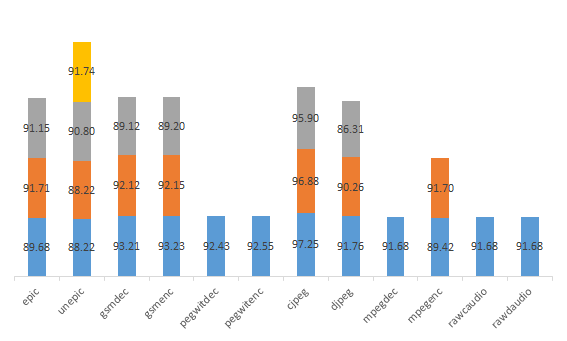}
  \caption{Pareto front energy consumption with respect to the baseline configuration (labels represent the percentage of improvement in energy consumption). Each color represents one point in the non-dominated front, as in Figure \ref{Figure3}.}
  \label{Figure4}
\end{figure*}

\begin{table}
\centering
\begin{tabular}{|ccc|}
\hline
\textbf{Application} & \textbf{Execution Time} & \textbf{Energy Consumption} \\ 
\hline
epic       & 62.79  & 90.85 \\ 
unepic     & 60.74  & 89.75 \\ 
gsmdec     & 63.12  & 91.48 \\ 
gsmenc     & 63.21  & 91.52 \\ 
pegwitdec  & 60.09 & 92.43 \\ 
pegwitenc  & 59.74 & 92.55 \\ 
cjpeg      & 87.23 & 96.68 \\ 
djpeg      & 62.37 & 89.44 \\ 
mpegdec    & 63.45 & 91.68 \\ 
mpegenc    & 63.28 & 90.56 \\ 
rawcaudio  & 63.59 & 91.68 \\ 
rawdaudio  & 63.58 & 91.68 \\ 
\hline
\textbf{Average}       & \textbf{64.43} & \textbf{91.69} \\
\hline
\end{tabular}
\caption{Percentage of improvement, averaged for each Pareto front and single objective vs. baseline cache configuration.}
\label{table_average_POS}
\end{table}

Figures~\ref{Figure3} and~\ref{Figure4} show the level of improvement computed for each point in the Pareto fronts obtained for all the 12 target applications. Figure~\ref{Figure3} depicts the percentage of improvement in execution time, whereas Figure~\ref{Figure4} depicts the level of improvement in energy consumption. As can be seen, our approach achieves a significant improvement in both objectives. In this regard, Table~\ref{table_average_POS} shows these improvements averaged over each Pareto front, and their averages in the last row. Our optimization method is able to reach cache configurations which are, in average, a 64.43\% and 91.69\% better in execution time and energy, respectively. To better understand this high improvement, we must compare the baseline configuration with, for example, the optimized cache configuration shown in Figure \ref{fig:ParetoSets}. Firstly, the baseline configuration has a 16 bytes block size, whereas the optimized configuration has an 8 bytes block size. Moving 16 bytes from main memory to cache memory consumes more energy than moving 8 bytes. Secondly, the baseline configuration has 64 ways versus the 4 ways of the optimized version. It means that the baseline configuration is much more associative and then finding the desired block spends much more time and energy, since each label must be compared 64 times against 4. Finally, the prefetch policy of the baseline instructions cache configuration is ``ON-DEMAND'', whereas in the optimized cache is ``ALWAYS''. Instructions are usually loaded from consecutive memory addresses (discarding branch instructions), and thus, the ``ON-DEMAND'' prefetch policy will consume more time and energy than the optimized ``ALWAYS'' prefetch policy \cite{Hennessy2011}.

Regarding the convergence of the optimization process, Table~\ref{table_sumarize} shows a summary of the evolution for both objectives. Column labeled as INI represents the level of improvement for each objective averaged over the initial random population. Column labeled as END represents the same values averaged over the final population. Column AVG shows averaged improvements for each objective and over all the generations and individuals, from INI to END. As Table \ref{table_sumarize} shows, NSGA-II easily improves the performance of the baseline cache even after the first generation. The same does not happen in energy, where after the first generation, only in the case of cjpeg NSGA-II is able to improve the energy consumption of the baseline cache. Fortunately, after 3-4 generations, NSGA-II quickly find cache configurations that improve the baseline cache in both performance and energy. Surprisingly, the improvement in energy, which started with worst values, quickly grows up and reach much better values than the improvement in execution time (up to $97\%$ in the case of cjpeg). In summary, Table \ref{table_sumarize} demonstrates that our optimization methodology, even when starting from bad initial solutions, is able to reach high levels of improvement with respect to a baseline configuration.

\begin{table}
\centering
\begin{tabular}{|ccccccc|}
\hline
  \multirow{2}{*}{\textbf{Application}}&\multicolumn{3}{|c|}{\textbf{Execution Time}} &\multicolumn{3}{|c|}{\textbf{Energy Consumption}} \\
   \cline{2-7} & \multicolumn{1}{|c}{\textbf{INI}} & \textbf{AVG} & \textbf{END} & \multicolumn{1}{|c}{\textbf{INI}} & \textbf{AVG} & \textbf{END}\\ 
\hline 
epic       & 11.40 & 57.04 & 63.11 & -71.42 & 77.56 & 91.71\\ 
unepic     & 12.55 & 54.97 & 61.80 & -95.27 & 75.79 & 91.74\\ 
cjpeg      & 66.13 & 85.01 & 87.51 & 48.03  & 91.77 & 97.25\\ 
djpeg      & 12.88 & 55.26 & 62.77 & -69.11 & 74.77 & 91.91\\ 
gsmdec     & 1.25 & 53.37 & 63.52 & -97.14  & 77,22 & 93,19\\ 
gsmenc     & 4.46  & 55.01 & 63.65 & -71.97 & 77.58 & 93.20\\ 
rawcaudio  & 12.63 & 57.27 & 63.59 & -80.69 & 76.90 & 91.68\\ 
rawdaudio  & 10.27 & 57.50 & 63.58 & -60.05 & 77.95 & 91.68\\ 
mpegdec    & 2.18  & 57.17 & 63.45 & -105.70 & 76.63 & 91.68\\ 
mpegenc    & 4.02  & 57.08 & 63.37 & -66.60  & 76.97 & 91.74\\ 
pegwitdec  & -1.02 & 51.56 & 60.09 & -146.23 & 73.00 & 92.46\\ 
pegwitenc  & 12.07 & 51.03 & 59.74 & -161.10  & 71.76 & 92.55\\ 
\hline
\end{tabular}
\caption{Improvement percentages: initial, average and final improvements per application vs. the baseline configuration.}
\label{table_sumarize}
\end{table}

\subsection{Validation with two additional baseline cache}

To validate our optimization framework, we have optimized the cache memory of two additional different hardware platforms included in some Apple devices of the family of SoC Apple AX, like iPhone 5, iPhone 5s, iPad 2, iPod-touch or iApple-TV. Apple AX series integrate the ARM processors family, for instance Cortex-A9 (iPad 2, iPod-touch or iApple-TV) or Cortex-A15 (iPhone 5, iPhone 5s). According to this, these two new cache configurations are:

\begin{itemize}
 \item[Baseline 2]: Cache size: 32 KB; Block size: 64; Associativity: 4; Replacement algorithm: RANDOM: Prefetch policy: ALWAYS; Write policy (DCache): COPY-BACK;
 \item[Baseline 3]: Cache size: 32 KB; Block size: 64; Associativity: 2; Replacement algorithm: LRU: Prefetch policy: ALWAYS; Write policy (DCache): COPY-BACK;
\end{itemize}

\begin{table}
\centering
\begin{tabular}{|c|cccc|}
\hline
\multirow{2}{*}{\textbf{Application}}&\multicolumn{2}{|c|}{\textbf{Baseline 2}} &\multicolumn{2}{|c|}{\textbf{Baseline 3}}\\
\cline{2-5} & \textbf{ExTime\%} & \textbf{Energy\%} & \textbf{ExTime\%} & \textbf{Energy\%} \\ 
\hline
epic&19.80&87.60&11.77&88.66\\
unepic&9.73&87.28&1.05&88.34\\
gsmdec&23.97&89.35&16.98&90.27\\
gsmenc&24.18&89.51&16.98&90.37\\
pegwitdec&34.26&92.25&28.12&92.72\\
pegwitenc&35.42&92.61&29.49&93.05\\
cjpeg&21.11&87.98&13.11&88.98\\
djpeg&27.38&88.74&21.01&89.93\\
mpegdec&24.80&88.01&18.61&89.42\\
mpegenc&22.20&87.65&14.32&88.70\\
rawcaudio&22.80&87.97&14.68&88.92\\
rawdaudio&24.13&87.81&16.14&88.77\\
\hline
\textbf{Average}&24.15&88.90&16.85&89.84\\
\hline
\end{tabular}
\caption{Percentage of improvement for the best cache configuration obtained vs. new baselines chosen.}
\label{tab:newBas}
\end{table}

We have repeated the process for the two additional baselines, computing the percentage of improvement obtained in the comparison of each baseline cache memory with the best cache configuration obtained by the optimization framework. Table \ref{tab:newBas} shows the percentages obtained, in execution time and energy consumption. It is worth noting that whereas the energy savings are still similar to Baseline 1 (close to 90\%), the improvement in execution time is decreased (from 64\% to 24\% and 17\%, respectively). These differences are given to the nature of the MediaBench benchmark and each specific baseline architecture. The first baseline, a Car GPS device, is oriented to a very specific navigation application, completely different in nature to MediaBench. It explains the high level of optimization in both execution time and energy. On the other hand, baselines 2 and 3 are general-purpose devices, and their cache memories are oriented to a wide range of different block sizes. It is translated into a low associativity but a big size, or in other words, better execution time but more energy, which explains the low level of improvement in execution time versus the high level of improvement in energy. 

\subsection{On the performance of the optimization framework}

Finally, with respect to the execution time of the optimization process, our master-worker architecture computed optimized cache configurations in an averaged wall-clock time of 10 hours (0.42 days) per application. The optimization of the set of 12 applications was performed in 5 days. Taking into account that the averaged time used by our simulation framework to evaluate one single cache configuration is equal to 13 seconds, an exhaustive optimization algorithm would take almost 10 days to find the Pareto optimal front for one single application, and 117 days to reach the set of 12 Pareto optimal fronts. As a result, a parallel master-worker NSGAII algorithm obtain excellent solutions (64.43\% and 91.69\% better in execution time and energy, respectively) with a difference of more than three months, obtaining a speed-up of 23.4 with respect to the exhaustive algorithm.

\section{Conclusion and future work}\label{sec:Conclusions}

Current multimedia embedded devices like smartphones, video players, etc. are highly constrained from battery lifetime and performance. Cache memories are added to these devices in order to improve performance. However, the selection of the best cache configuration for each embedded system is a hard task because of the large space of possible cache configurations. Several design techniques have been proposed for years in order to facilitate the search of the best cache configuration for different applications.

In this paper, we have presented a novel technique based on static profiling and multi-objective optimization to find the best cache configuration for a given target embedded system and a target set of applications. The process has been divided in two phases: the first one is responsible for obtaining the program traces and parameters needed to characterize the set of candidate cache configurations. The second phase applies multi-objective evolutionary algorithm, using NSGA-II and Dinero IV, to evaluate each application under the candidate set of cache configurations.

The result of the optimization is a set of cache configurations that minimizes execution time and energy consumption for each application. Therefore, this improves the performance and increases the lifetime of both batteries and devices. Taking a cache configuration commonly used in current multimedia systems as a baseline, experimental results show an average improvement of $64.43\%$ and $91.69\%$ in execution time and energy consumption, respectively.

Our methodology still needs human decisions to select the final cache memory, the best possible for the whole set of applications. We have seen that this is not a difficult task. However, as our future work, we are already extending this methodology to allow us the automatic optimization of all the target applications at a time. This will require a greater parallelization degree of the evaluation process and the design of a new accurate multi-objective function, incorporating for instance fuzzy decisions to reduce the number of objectives from $12~\mathrm{applications} \times 2~\mathrm{objectives}$ to two or three objectives.

\section*{Acknowledgment}
This work has been partly funded by the Spanish Ministry of Economy and Competitivity under research grants TIN2014-54806-R and TIN2014-56494-C4-2-P.

\bibliographystyle{elsarticle-num}
\bibliography{biblio}

\appendix
\section{Multi-objective optimization and hypervolume indicator}\label{sec:AppA}
\subsection{Multi-objective optimization}
Multi-objective optimization aims at simultaneously optimizing several contradictory objectives. For such kind of problems, a single optimal solution does not exist, and compromises have to be made. Thus, without any loss of generality, we can assume the following formulation of the m-objective minimization problem:

\begin{eqnarray} 
\mathrm{Minimize} & & \nonumber \\
\mathbf{y} = \mathbf{f}(\mathbf{x}) & = & [f_1(\mathbf{x}), f_2(\mathbf{x}), . . . , f_m(\mathbf{x})] \nonumber \\
\mathrm{Subject~to} & & \nonumber \\
\mathbf{x} & = & (x_{1}, x_{2}, \ldots , x_{n}) \in \mathbf{X} \nonumber \\
                    \mathbf{y} & = & (y_{1}, y_{2}, \ldots , y_{m}) \in \mathbf{Y} \nonumber
\end{eqnarray}
 
where $\mathbf{x}$ is the vector of $n$ decision variables, $\mathbf{f}$ is the vector of $m$ objectives function. $\mathbf{X}$ is the feasible region in the decision space, and $\mathbf{Y}$ is the feasible region in the objectives space. A solution $\mathbf{x_1} \in \mathbf{X}$ is said to dominate another solution $\mathbf{x_2} \in \mathbf{X}$ (denoted as $\mathbf{x_1} \prec \mathbf{x_2}$) if the following two conditions are satisfied:

\begin{eqnarray}  
\forall i \in {1,2,....,m} & , & f_i(\mathbf{x_1}) \leq f_i(\mathbf{x_2}) \nonumber \\
\exists j \in {1,2,....,m} & , & f_j(\mathbf{x_1}) < f_j(\mathbf{x_2}) \nonumber 
\end{eqnarray}

If there is no solution which dominates $\mathbf{x} \in \mathbf{X}$, $\mathbf{x}$ is said to be a non-dominated solution. The non-dominated set of the entire feasible search space $\mathbf{X}$ is the \emph{Pareto optimal set}. The image of the Pareto optimal set in the objective space is the \emph{Pareto optimal front} of the multi-objective problem at hand. Since several solutions may be mapped to the same multi-objective function, the Pareto optimal front does not necessarily contain as many elements as the Pareto optimal set. A multi-objective optimization problem is solved, when its complete Pareto optimal set is found. In practice, the number of Pareto optima is too large, or the determination of a single Pareto optimum is NP hard \cite{Deb2009}. Therefore, the aim is usually to find a satisfactory Pareto set approximation (usually named Pareto set, or Pareto front in the objective space), as close as possible to the Pareto optimal set.

\begin{figure}[ht]
  \centering
  \includegraphics[width=0.45\textwidth]{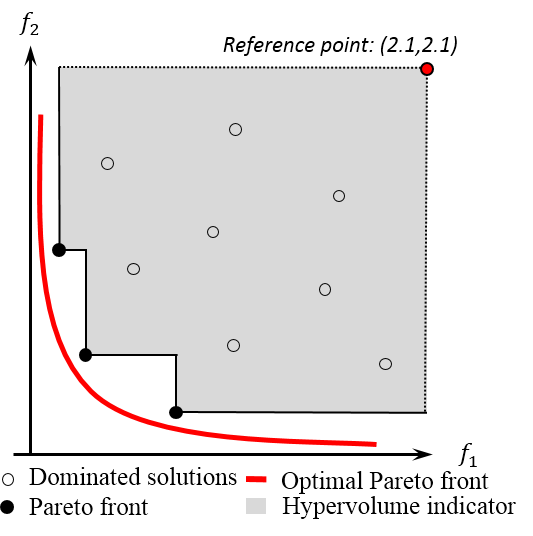}
  \caption{Illustration, in a two objective space, of the concepts of dominance, Pareto optimal front, Pareto front, and hypervolume indicator. For the last one, the objective vector (2.1,2.1) is taken as the reference point.}
  \label{fig:ParetoFrontExplanation}
\end{figure}
 
Figure \ref{fig:ParetoFrontExplanation} depicts an example of a Pareto optimal front (continuous curve), the set of solutions obtained by a given optimization algorithm (black shaded and non-shaded circles), the subset of non-dominated solutions (black shaded circles) that form a Pareto front, and the subset of dominated solutions (non-shaded circles). In the following, we show how to analyze the quality of a obtained Pareto front using the hypervolume indicator.

\subsection{Hypervolume indicator}
The hypervolume indicator is a metric that calculates the volume (in the objective space) covered by members of a non-dominated set of solutions $Q$ \cite{Deb2009}. Let $v_i$ be the volume enclosed by solution $i \in Q$. Then, a union of all hypercubes is found and its hypervolume ($I_H$) is calculated as:

\begin{equation}
I_H(Q) = \bigcup_1^{\left| Q \right|} v_i
\end{equation}

The hypervolume of a set is measured relative to a reference point, usually the anti-optimal point or ``worst possible'' point in space. We do not address here the problem of choosing a reference point. If the anti-optimal point is not known or does not exist one suggestion is to take, in each objective, the worst value from any of the fronts being compared. In this work, we consider the hypervolume difference to a reference set $R$, defined as
\begin{equation}
I_H^-(Q) = I_H(R) - I_H(Q)
\end{equation}
where smaller values correspond to higher quality. Since the reference set is not given, we take $I_H(R) = 0$. 

Figure \ref{fig:ParetoFrontExplanation} shows an example of how $I_H(Q)$ is computed using the reference point $(2.1, 2.1)$. 

\end{document}